Smiling Women Pitching Down:

Auditing Representational and Presentational Gender Biases in Image Generative AI

Luhang Sun[1], Mian Wei[1], Yibing Sun[1], Yoo Ji Suh[1], Liwei Shen[2], and Sijia Yang[1]

[1] School of Journalism and Mass Communication, University of Wisconsin–Madison
[2] Department of Communication Arts, University of Wisconsin–Madison

Corresponding author: Sijia Yang; e-mail: syang84@wisc.edu

**Abstract**

Generative AI models like DALL·E 2 can interpret textual prompts and generate high-quality images exhibiting human creativity. Though public enthusiasm is booming, systematic auditing of potential gender biases in AI-generated images remains scarce. We addressed this gap by examining the prevalence of two occupational gender biases (representational and presentational biases) in 15,300 DALL·E 2 images spanning 153 occupations, and assessed potential bias amplification by benchmarking against 2021 census labor statistics and Google Images. Our findings reveal that DALL·E 2 underrepresents women in male-dominated fields while overrepresenting them in female-dominated occupations. Additionally, DALL·E 2 images tend to depict more women than men with smiling faces and downward-pitching heads, particularly in female-dominated (vs. male-dominated) occupations. Our computational algorithm auditing study demonstrates more pronounced representational and presentational biases in DALL·E 2 compared to Google Images and calls for feminist interventions to prevent such bias-laden AI-generated images to feedback into the media ecology.

**Keywords**

Generative AI, DALL·E 2, gender bias, algorithm auditing, computer vision, representational bias, presentational bias

**Introduction**

The advent of generative deep learning models, such as large-scale language models (LLMs) and Diffusion Models, has brought Generative Artificial Intelligence (Generative AI) to the spotlight of public attention. OpenAI, the parent company behind ChatGPT, released DALL·E 2 in 2022, a text-to-image generative AI product that allows users to generate images using descriptive textual prompts. As impressive as DALL·E 2's capacity to produce high-quality, multi-style (e.g., photorealistic, artistic) and impressive images exhibiting human creativity (e.g., blending multiple concepts in the textual prompt), the risk for algorithmic bias, particularly the potential to reinforce and amplify gender bias, cannot be overlooked.

In a report by OpenAI (2022), the company acknowledges that DALL·E 2 tends to generate more images of men than women when given gender-neutral prompts and notes that filtering training data may have intensified such biases. Due to the lack of transparency, both the algorithms and the training materials of this generative AI product remain a "black box." Consequently, concerns have been raised about the ethical implications of generative AI's potential to reproduce and exacerbate gender biases in today's media ecology—unlike previous media technologies that primarily influence content selection, filtering, and curation (e.g., Google Images, Facebook's newsfeed recommendation algorithms), generative AIs are unique in their capacity to directly participate in content creation (Quadflieg et al., 2022). Therefore, empirical evidence is urgently needed to audit the presence, magnitude, and types of gender biases in generative AIs such as DALL·E 2 before its widespread adoption systematically skews the visual media landscape along gender lines.

It is essential to note that gender bias is not merely a matter of counting women and men in AI-generated images. Although prior research auditing algorithmic gender biases (e.g., Google images search) typically focus on the unequal representation of men versus women across settings and roles such as occupations (i.e., representational bias, see Kay et al., 2015; Lam et al., 2018), there is a longstanding literature documenting stereotyped media presentation of men versus women (e.g., smiling women and calm men, Grau & Zotos, 2016). Such presentational biases have received less attention from scholars studying gender biases in algorithmic communication. We aim to empirically document the prevalence and magnitude of both types of gender biases in AI-generated images, with DALL·E 2 as a case study.

In the following sections, we first review the literature on gender biases and stereotypes in algorithms and emerging AI technologies, and unpack representational versus presentational gender bias in mass media. Then, we extend the application of these constructs from the mass media context to generative AI and, following prior literature (Kay et al., 2015; Lam et al., 2018), focus on occupational gender biases, given the availability of census labor statistics as a benchmark. We compiled a list of occupations from the 2021 US Current Population Survey (CPS) census data and developed textual prompts accordingly to gather image data from DALL·E 2 and Google Images, respectively. Finally, we computationally extracted visual features (e.g., human faces, facial expressions, calmness, head pitching) from DALL·E 2 images and compared them with benchmark data (i.e., the 2021 census labor statistics and Google Images data). Our results confirmed the prevalence of both representational and presentational gender biases in AI-generated images. Importantly, we found evidence demonstrating bias amplification in DALL·E 2 relative to benchmark data, while Google Images exhibits signs of bias countering. Our findings suggest the need for continued auditing of rapidly evolving generative AI technologies and for feminist interventions to prevent such bias-laden AI-generated visuals from permeating into the current media ecology already fraught with gender biases.

**Literature review**

*Algorithmic biases, generative AI, and gender bias amplification*

Algorithms have become ubiquitous in contemporary society, playing a pervasive role in the automation of an extensive range of tasks. However, systematic and repeatable algorithmic bias can arise during any stage of algorithms' permeation into everyday life, including data processing, algorithm development, and the interaction between algorithms and humans (Olteanu et al., 2019; Mehrabi et al., 2021; Suresh & Guttag, 2021). In a study on the influences of algorithms and data, Quadflieg et al. (2022) argue that the impact of algorithmic power is not evenly distributed, but instead amplifies existing power disparities. Previous research has shown that algorithms can encode biases related to race, sexuality, and social class into daily life. For instance, healthcare algorithms have been found to exhibit racial bias, preventing Black patients from receiving the same medical services as white patients (Obermeyer et al., 2019). Monea (2022) argues that the internet becomes straight as it suppresses LGBT-related content through opaque algorithmic filters. Similarly, algorithms used in recruitment and pedestrian detection have been shown to discriminate against women (Cappelli et al., 2018; Brandao, 2019). With the increasing use of algorithmic and AI products, there is a risk that they may threaten the progress made in global equality and human rights (Bartoletti, 2020).

  One of the most significant gender biases embedded within algorithmic systems is the stereotyped occupational roles of men and women. For example, Google translates the Turkish phrase "O bir doktor" as "he is a doctor" in English, but the translation becomes "she is a nurse" if the word "doctor" is replaced by the Turkish word "nurse" even though the Turkish pronouns are gender neutral (Johnston, 2017). Word embeddings link "she" to occupations such as homemaker, nurse, and receptionist while associating "he" with philosopher, architect, and financier (Bolukbasi et al., 2016). Additionally, digital assistants like Siri often have female voices by default, which reinforces implicit gender prejudices expecting women to play assisting roles (LaFrance, 2016).

  Prior research on search and recommender algorithms has indicated that visual representations of women and men not only reflect but can even amplify existing stereotypes. For instance, image search results for occupations tend to systematically underrepresent women and exaggerate gendered stereotypes (Kay et al., 2015): women are underrepresented in image search results across 57% of 105 jobs (Lam et al., 2018), and search engines display much more images of men than women when searching for "CEO" online (Quadflieg et al., 2022). Stereotypes of gendered occupations persist across different digital platforms, including Wikipedia and Shutterstock (Singh et al., 2020). Recommendation systems for job ads also contribute to the reinforcement of traditional beliefs of gender roles and encourage the gendered division of labor, further widening the gender gap in the digital space and perpetuating gender bias in society (Gibbs, 2015; Lambrecht & Tucker, 2019; Wood & Eagly, 2012).

  Generative AI, including ChatGPT, GPT-4, DALL·E 2, and Midjourney, has become increasingly popular. The breakthrough in large-scale language models and deep learning allows generative AI to produce multimodal content in several seconds (Lawton, 2023). Currently, more than three million people use DALL·E 2 to produce over four million images daily (Wiggers, 2022). However, the rise of generative AI also raises ethical concerns, as it risks exacerbating gender biases in the digital space to a greater degree than other AI products. OpenAI (2022) acknowledges that gender bias exists in their AI-generated images and attributes the bias partially to "images from the internet." Compared with online search tools such as Google

Images that also exhibit gender bias (Kay et al., 2015; Lam et al., 2018), generative AI systems are potentially a game changer, as they directly participate in content production and thus risk pumping bias-infused content back into the media ecology. Upon widespread adoption, such AI-generated content may serve as biased training data for other generative AI products or future iterations of the same AI, if no screening is applied. This would create a vicious feedback loop to reproduce and reinforce gender biases. With generative AI products updating so rapidly, it is essential to examine the prevalence, magnitude, and types of gender biases in popular tools such as DALL·E 2 to alert researchers, developers, policymakers, and the public in a timely way.

Existing studies auditing gender biases in generative AI technologies tend to focus on textual output (Kirk et al., 2021; Lucy & Bamman, 2021). For instance, an empirical analysis of GPT-2 reveals that machine-predicted occupations are more stereotypical and less diverse for women (Kirk et al., 2021). Stories generated by GPT-3 tend to associate feminine characters with domestic roles and physical appearances while describing them as less powerful than their masculine counterparts (Lucy & Bamman, 2021). Similar gender bias auditing studies on image generative AI technologies such as DALL·E 2 are still lacking. Given that generative AI is a relatively new domain for study in communication research, we first review relevant literature on gender representation and biases in mass media and digital communication technologies.

*Representational gender bias in mass media and its negative effects*

Mass media have long been critiqued for inaccurate and stereotypical representations of reality (Noelle-Neumann, 1993; Seiter, 2006), perpetuated by powerful actors who invest significant resources in maintaining the *status quo* to serve their interests, including patriarchy, heterosexism, and capitalism (Entman, 2007). Among these biases, unequal and stereotypical media portrayals of men and women are not merely reflections of existing inequalities but rather active practices that exacerbate gender oppression (Shor et al., 2015).

Previous research has shown that gender biases in media can harm women in two ways. First, negative portrayals of women affect their self-perception and cognitive and educational achievements. Through a meta-analysis of 33 experiments, Appel and Weber (2021) found that devaluing media content impaired the cognitive and educational achievement of members of the stereotyped groups. In contrast, nonmembers were not affected or even benefited from such biased media content. An experimental study shows that gender-stereotypical television commercials restrain women's performance in math and choices of career path (Davies et al., 2002). Second, media content instills harmful stereotypes about women into other people's minds. For example, media consumption fosters male college students' beliefs in gender stereotypes about black women such as Jezebel and Sapphire (Jerald et al., 2017). Such gender biases are prevalent in various visual media, including print images (Parker et al., 2017), television dramas and commercials (McArthur & Resko, 1975; Parker et al., 2017), and social media images (Döring et al., 2016).

In reviewing the literature, two types of media biases regarding the visual portrayal of women become prominent: *representational bias* and *presentational bias*. Regarding representational bias, on the one hand, women have been traditionally underrepresented over men across visual media. Based on the quantitative content analyses of the two special issues of *Sex Roles*, Collins (2011) discovered that half of the empirical articles (nine of 18) examining gender roles in media find that women are portrayed less frequently in at least one content category. In addition, male main characters appear nearly twice as female characters in 200 award-winning children's picture books. Male characters outnumber female characters by 53% in the illustrations (Hamilton et al., 2006). Regarding occupational representations, women are less likely to be depicted as having professional or science jobs (Coltrane & Adams, 1997; Kerkhoven et al., 2016).

On the other hand, mass media excessively associate women with domestic and stereotypical occupational roles. Television advertising usually presents women being at home in dependent roles (Knoll et al., 2011). Women in the workplace tend to be represented as nonprofessionals, homemakers, and sexual gatekeepers, occupying positions without authority or even without pay (Collins, 2011; Hamilton et al., 2006). In addition, women are more likely to be employed in service, clerical, or teaching occupations (Coltrane & Adams, 1997; Kerkhoven et al., 2016). Representational bias against women in mass media is harmful to women's confidence and performance improvement. Suppressed media visibility may reinforce entrenched status beliefs, signaling that women are not seen as equally competent and important as men (Shor et al., 2015). Good and colleagues (2010) also found that images featuring male scientists impaired female students' science performance, while exposure to counter-stereotypical images (e.g., female scientists) improved their comprehension. Therefore, We define *representational bias* as the unequal representation of men versus women across various media settings and particularly the overrepresentation of women in stereotypically feminine roles.

Given the rise of visual media, recent studies have begun to examine representational gender bias in online images portraying men versus women across occupations (Lam et al., 2018; Kay et al., 2015). These studies confirmed that representational bias does exist in online images searched through Google: overall, women were underrepresented compared to men across occupations; and importantly, for more than half the audited occupations, women were more underrepresented relative to their actual participation in these jobs. Notably, this representation deficit between the share of women in online images and their actual share in the workforce was even more severe for occupations already dominated by men, thus demonstrating the amplification of representational bias by Google Images. Following the burgeoning research auditing algorithmic gender biases, we aim to estimate the prevalence and magnitude of representational bias in DALL·E 2 images. To benchmark against existing gender disparities in labor statistics and evaluate potential bias amplification effects of DALL·E 2, we followed previous research (Kay et al., 2015) and categorized occupations into *male-dominated*, *female-dominated*, and *relatively-equal* jobs (see Methods for operationalization details). We expect DALL·E 2 to exacerbate existing gender segregation in census labor statistics. More importantly, given that DALL·E 2 was trained on online images and that its model development process may have further entrenched gender biases (OpenAI, 2022), we expect DALL·E 2's amplification of representational bias will be more severe than Google Images.

>*H1*: DALL·E 2 tends to underrepresent women in male-dominated occupations and overrepresent women in female-dominated occupations than in census data.
>
>*H2*: DALL·E 2 tends to underrepresent women in male-dominated occupations and overrepresent women in female-dominated occupations than Google Images.

*Presentational gender biases: Emotions and Gestures*

Compared to *representational bias,* which pertains to the unequal distribution of men and women in aspects such as roles, abilities, behaviors, and occupations, *presentational bias* focuses on how media portray individuals differently based on their gender, often by highlighting certain emotions, gestures, traits, or physical characteristics that are seen as stereotypically male or female. Research has documented biased media presentations of women concerning their emotions, traits, and color attributes (Grau & Zotos, 2016). For instance, women are frequently portrayed as happy and smiling in news photographs, reflecting cultural expectations of women to behave in a positive and "lady-like" manner (Rodgers et al., 2007). Female politicians are also more likely to display visible positive emotions on television compared to male politicians (Renner & Masch, 2019). This gendered pattern can be attributed to the long-standing stereotype that women experience and express more emotions while men are calm and rational (Plant et al., 2000). As a result, female leaders can be criticized for showing even minor negative emotions or emotions that convey dominance such as anger and pride. However, unemotional women can also be penalized for not fulfilling their warm and communal roles (Brescoll, 2016).

In addition to emotional biases, female images often exhibit subordination and passivity through facial expressions and body gestures (Collins, 2011; Grau & Zotos, 2016), while men are often portrayed as dominant and confident (Plakoyiannaki et al., 2008). Such stereotypes attribute greater competence and status worthiness to men and can contribute to the "glass ceiling" phenomenon that prevents women from assuming leadership roles. They also legitimize penalizing assertive women leaders for violating gender hierarchy (Ridgeway, 2001). Moreover, facial orientation can affect people's perceptions of power. Faces pitched upward (low camera angle) convey a sense of authority and dominance compared with faces pitched downward (high camera angle) (Grabe & Bucy, 2009; Peng, 2018). Because media tend to present dominant men and subordinated women, male and female characters may display different face-pitching angles. Figure 1 shows examples of images generated from DALL·E 2 that feature presentational biases of emotions and face-pitching.

[Figure 1 about here]

Building upon extensive documentation of *presentational biases* in mass media portrayals of women, we examine the presence and magnitude of presentational biases regarding smile, calmness, and face-pitching in DALL·E 2 generated images. Although previous research has applied computer vision techniques to compare how male versus female politicians were stereotypically portrayed in news images along gender lines (Peng, 2018), potential presentational biases in generative AI have not yet received much attention. We aim to fill this gap by first auditing whether DALL·E 2 tends to amplify presentational biases by occupation. For instance, if women in DALL·E 2 images tend to smile more than men, and if further this stereotypical portrayal of smiling women is particularly pronounced for occupations already with female overrepresentation based on census labor force statistics, we consider this evidence supporting DALL·E 2's *amplification* of the presentational bias of smiling. Given that DALL·E 2 sources training data from the internet (OpenAI, 2022), we further compare whether such bias amplification is more severe in DALL·E 2 than in Google Images, the most widely used search tool for online images. Benchmarking against Google Images can help assess whether generative AI technologies such as DALL·E 2 may pose additional risks for exacerbating presentational gender biases.

> *H3*: (a) In DALL·E 2 images, women will be more likely to *smile* than men especially in female-dominated occupations, and (b) such amplification of presentational bias regarding *smile* will be more pronounced in DALL·E 2 than Google Images.
>
> *H4*: (a) In DALL·E 2 images, women will be less likely to show *calmness* than men especially in female-dominated occupations, and (b) such amplification of presentational bias regarding *calmness* will be more pronounced in DALL·E 2 than Google Images.
>
> *H5*: (a) In DALL·E 2 images, women will be more likely to *pitch downward* than men especially in female-dominated occupations, and (b) such amplification of presentational bias regarding *pitch* will be more pronounced in DALL·E 2 than Google Images.

**Methods**

*Datasets*

Three datasets were assembled for analysis: a) 2021 Current Population Survey (CPS) census data on occupational gender segregation, b) Google Images data by occupation, and c) generative AI images by occupation. The first two datasets serve as the benchmark, because the former provides information on current occupational gender disparities while the latter represents the most common source of online images. We aim to evaluate the prevalence, magnitude, and types of gender biases in generative AI images, obtained from the DALL·E 2 image generation API endpoint, against each of these benchmark datasets.

*2021 CPS census data* were collected annually and released by the Bureau of Census for the Bureau of Labor Statistics (BLS) in the United States. The 2021 CPS census data report weekly income and percentages of men versus women currently employed in a total of 22 broader occupational categories and 565 occupations. We used these data to benchmark the prevalence of existing gender disparity[2] by occupation. In data preprocessing, we first filtered out occupations without information on gender disparity, reducing the initial dataset to 354 occupations. Then, for each occupational category, we selected the top 50% of occupations with the largest number of employees. We further pilot tested prompting DALL·E 2 to generate images, occupation by occupation, and removed occupations yielding an insufficient number of images with detectable human faces. This preprocessing process led to a finalized list of 153 occupations (See Supplemental Materials for details on the screening procedure).

*Google Images data* serve as our second benchmark, collected via an online scraping tool SerpAPI.[3] We created textual search terms for each occupation on the finalized list of the 2021 CPS census data and collected 100 Google images per occupation ($N_{Google}$ = 15,300).

*Generative AI images* are obtained from the DALL·E 2 image generation API endpoint from OpenAI. This dataset similarly contains 100 images for each occupation ($N_{DALL·E\ 2}$ = 15,300) as the Google Images dataset. To exclude images without human faces, we used Amazon AWS *Rekognition*[4] for face detection, as further detailed in the next section. We continued to collect images from both DALL·E 2 and Google Images until we obtained 100 images with detectable human faces for each of the 153 occupations on our finalized list.

## Measures

To extract visual features from collected images, we utilized Amazon AWS *Rekognition* that enables us to efficiently process our large image corpus and assemble an analytical dataset with detailed image-level visual features. Amazon AWS *Rekognition* uses deep learning and computer vision algorithms to annotate images and extract visual features, including human face detection, gender detection, and emotion recognition.

Amazon AWS *Rekognition* detects whether an image contains any human face, the number of faces, and facial features including gender (binary, male or female), smile (binary, yes or no), emotions (eight types of discrete emotions), and pose (yaw, pitch, and roll). *Rekognition* identifies facial landmarks such as eyebrows and mouth and draws bounding boxes around the detected faces. We further validated machine-coded faces, gender, smile, and emotions against human annotations provided by three undergraduate coders blind to study hypotheses (average Krippendorff's alpha = .90, via the *ReCal* web service, see Freelon, 2013): the average Precision, Recall, and F-score are .85, .92, and .87, respectively (See Supplemental Materials for detailed information on human validation).

In many images, multiple faces coexist. Since the largest face typically grabs the most attention (Min et al., 2017), we focused our analyses on the most prominent face in such multi-face images. We calculated the area of each bounding box enclosing a detected face and selected the face with the largest area size.

Furthermore, to better assess how DALL·E 2 may amplify existing occupational gender biases relative to Google Images, we categorized the 153 occupations in the 2021 CPS dataset into three groups based on the percentages of female employees: *male-dominated* occupations were defined as those employing less than 33.3% females ($N = 57$), *female-dominated* occupations employing more than 66.7% females ($N = 44$), and relatively-equal occupations employing between 33.4% and 66.6% females ($N = 52$).

Statistical Analysis

To investigate representational gender bias in DALL·E 2, we conducted one-proportion Z-tests to compare, occupation by occupation, the proportion of females in DALL·E 2 images with the known proportion in the 2021 CPS census data (H1). Then, we carried out two-proportion Z-tests to compare each occupation-specific proportion of females in DALL·E 2 images to the corresponding proportion in the Google Images dataset (H2), treating both proportions as estimated quantities with inference uncertainties.

To examine presentational gender biases, we fitted generalized linear mixed models (GLMM) using maximum likelihood estimation (MLE) and Laplace approximation to predict the presence of a smiling face in each image, and linear mixed models (LMM) using restricted maximum likelihood (REML) estimation for continuous outcomes including calmness and pose pitch scores. We used type 3 Wald $\chi^2$ tests to test statistical significance in GLMM models, and $F$-statistics with Kenward-Roger degrees of freedom approximation for LMM models. All these models[6] included occupation types as random intercepts to account for the multilevel data structure where images (Level-1) were nested under occupation types (Level-2). In each multilevel regression model, we tested the fixed effects of three factors—gender (female vs. male), occupation types (two dummies, female-dominated vs. male-dominated, relatively-equal vs. male-dominated), and source (Google Images vs. DALL·E 2)—and their two-way and three-way interactions. For the source factor, we set DALL·E 2 = 0 as the reference group to obtain both conditional two-way interactions (gender × occupation types specific to DALL·E 2 images, H3a, H4a, and H5a) and three-way interactions (gender × occupation types × source, assessing how bias amplification further differed by source, H3b, H4b, and H5b) directly from the same multilevel regression model. For each presentational bias, we also estimated the degree of gender disparity across occupation categories for DALL·E 2 images (simple main effects of gender conditioned on the reference group) and how such gender disparity further differed by source (gender × source). All detailed results from multilevel regression analyses are presented in Appendix A.

## Results

### *Representational gender biases in DALL·E 2: systemic underrepresentation and stereotypical overrepresentation*

Our results showed systemic underrepresentation and stereotypical overrepresentation of women in DALL·E 2-generated images. Out of the 30,600 images with detected faces collected from both Google and DALL·E 2, 42.4% (12,983) were female and 57.5% (17,617) were male. Among the 15,300 Google images, 46.4% (7,105) were female and 53.6% (8,195) were male. Among the 15,300 DALL·E 2 images, 38.4% (5,878) were female and 61.6% (9,422) were male.

To test H1 and H2, we conducted a proportion Z-test to estimate the differences in female percentage comparing (a) DALL·E 2 images to the 2021 CPS census data, (b) Google images to the 2021 CPS census data, and lastly (c) DALL·E 2 images to Google images. The estimated differences were calculated by subtracting the female percentage in the census data from the female percentage in DALL·E 2 images (or Google images). As shown in Figure 2 Panel (a), there was representational gender bias in images generated by DALL·E 2. For the majority of male-dominated (e.g., architect and CEO, colored in blue) and relatively-equal occupations (e.g., writer and lawyer, colored in gray), women were significantly underrepresented in DALL·E 2 images. In contrast, for female-dominated occupations (e.g., cashier and nurse, colored in red), women were significantly overrepresented, which may reinforce occupational gender segregation unfavorable to women.

[Figure 2 and Table 1 about here]

Surprisingly, Figure 2 Panel (b) shows that Google images seem to counteract representational biases, returning images with higher proportions of men in female-dominated occupations and higher proportions of women in male-dominated occupations. The estimated differences in the share of female faces between DALL·E 2 images and Google images, shown in Panel (c), also confirms the prevalence of representational biases in DALL·E 2 images benchmarked against Google images: DALL·E 2 overrepresented women in female-dominated occupations while underrepresenting women in male-dominated and relatively-equal occupations. A more detailed visualization of the three comparisons, with 95% CIs quantifying inference uncertainties occupation by occupation, is presented in Figure 2 d-f. Based on the proportion Z-test results, both H1 and H2 are supported.

*Presentational gender biases in DALL·E 2*

*Smile*

To examine the presentational gender bias, GLMMs were fitted to predict the probability of an image containing a smiling face by gender, image source, and occupation category (Table 2 and Figure 3). The results indicate that across occupation categories, women were more likely to smile than men in DALL·E 2 images, $OR = 2.19$, 95% CI = [2.01, 2.39], $p < .001$; and further, this gender disparity was more severe than in Google images, $OR = 0.81$, 95% CI = [0.73, 0.91], $W(1) = 13.84$, $p = .001$. Regarding bias amplification, the conditional two-way interactions were significant for both *Gender × Relatively Equal*, $OR = 2.08$, 95% CI = [1.60, 2.71], $p < .001$; and *Gender × Female Dominated*, $OR = 2.50$, 95% CI = [1.87, 3.32], $p < .001$, suggesting that smiling faces were least likely to be present in male-dominated jobs. H3a was supported. Furthermore, the three-way interaction between gender, image source, and occupation category was also statistically significant, $W(2) = 14.95$, $p = .001$, suggesting that the severity of bias amplification varied by source. Figure 3b shows that detected gender bias amplification, operationalized as higher proportions of smiling women (vs. men) in female-dominated and relatively-equal occupations relative to male-dominated jobs, was more pronounced in DALL·E 2 images than Google images. Therefore, H3b was supported.

[Table 2 and Figure 3 about here]

*Calmness*

Next, we fitted LMMs with calmness scores as the outcome variable. Although H3 focused on smile, arguably highly correlated with positive emotions, we decided to examine calmness as well, as it can be viewed as a "neutral" or "lack of emotion." The results (Table 3 and Figure 4) showed that men were more likely to display calmness than women across occupation categories in DALL·E 2 images, $b = –11.94$, $p < .001$; and this gender disparity was more severe than Google images, $b = 3.45$, $F(1, 30567.4) = 15.18$, $p = .001$. Moving on to bias amplification, in DALL·E 2 images, neither of the two conditional two-way interactions comparing female-dominated/relatively-equal jobs to male-dominated jobs was significant. H4a was not supported. Furthermore, the three-way interaction between gender, occupation category, and image source was not statistically significant, $F(2, 27580.7) = 0.11$, $p = .899$, as seen in

Table 3 Model 2. This suggests the magnitude of bias amplification did not significantly differ between DALL·E 2 and Google images. Therefore, H4b was not supported.

[Table 3 and Figure 4 about here]

*Pose pitch*

Overall, we found greater variances in pitch scores compared to other features. Pitching downward typically signifies obedience and subordination. Surprisingly, there was no significant difference between men and women in the degree of downward pitching among DALL·E 2 images, and this pattern did not vary by image source. That said, the conditional two-way interaction *Gender × Female Dominated* was statistically significant ($b = –3.60, p = .023$), suggesting that within DALL·E 2 images, women exhibited a stronger tendency to pitch downward more than men, particularly in female-dominated jobs as compared to male-dominated jobs. H5a was supported. No evidence was found to support the three-way interaction, and therefore, H5b was rejected.

[Table 4 and Figure 5 about here]

**Discussion**

Building upon previous algorithm auditing research that has documented the prevalence of gender biases in Google Images (Kay et al., 2015; Lam et al., 2018) and the literature on mass and social media gender biases (Döring et al., 2016; McArthur & Resko, 1975; Parker et al., 2017), we employed a computational approach to empirically examine occupational gender biases in DALL·E 2, an increasingly popular image generative AI model released by OpenAI. Generative AI models such as DALL·E 2 hold the potential to revolutionize media content production, thereby posing a significant risk of reshaping today's media ecology in biased ways upon unscrutinized widespread adoption. After systematically comparing DALL·E 2 with Google Images and the 2021 census labor statistics across 153 occupations, we found evidence that DALL·E 2 risks amplifying both representational and presentational gender biases. Given the lack of transparency of model training and development process behind DALL·E 2, these findings underscore the importance to continuously monitoring gender biases in generative AI technologies through collaborative efforts by researchers, the industry, regulators, and the public.

First, DALL·E 2 systematically underrepresented women in male-dominated jobs while overrepresented women in images portraying female-dominated occupations. This is consistent with prior research that has documented similar representational gender bias in Google Images (Kay et al., 2015; Lam et al., 2018), with one important deviation: in our study, DALL·E 2 exhibited more severe representational bias than Google Image, which, in contrast, appeared to counter, not exacerbate, existing occupational gender segregation. We speculate that Google may have adjusted and improved its image search and recommendation algorithms to mitigate representational bias. Ramesh and colleagues (2021) reported that DALL·E 2 used "text-image pairs from the internet" (p. 4) for training but did not provide details on how the training dataset was constructed. This lack of transparency persists in the recent publication of the "GPT-4 Technical Report" by OpenAI (2023), in which OpenAI attributes biases present in DALL·E 2 generated images to existing biases within their current training data.7 However, our results revealed that Google, regarded as the most commonly used online image search engine, displayed less or even countering representational bias, with fewer instances of underrepresentation and overrepresentation of women across most occupations. Given the striking differences in representational bias between DALL·E 2 and Google Images documented in our study, it seems that DALL·E 2's biases cannot be attributed solely to online image data source similar to Google Images. The origin of representational gender bias seems to go beyond using training data of "images from the internet."

Since model development details of generative AI remain a "black box" with limited transparency, DALL·E 2 included, it is difficult to trace down and mitigate biases. Therefore, the booming generative AI industry should seek to establish a collaboration protocol with the academic community for data sharing, model performance auditing, and algorithm debiasing in a safe, commercially sound, and socially responsible way. Only by increasing the transparency of AI products can the public, AI professionals, and researchers from different fields participate in the collective decision-making processes to prevent the monopoly on AI technologies of several powerful stakeholders while reversing the tendency to exacerbate existing gender biases.

Second, we went beyond representational bias, the typical focus of prior research on algorithmic gender biases, to examine presentational biases including smiling, calmness, and pitching, based on research on mass and social media gender biases (Collins, 2011; Grau & Zotos, 2016; Peng, 2018; Renner & Masch, 2019; Rodgers et al., 2007). We found that across occupation categories, DALL·E 2 produced more images portraying smiling women and emotion-less, calm men, and such gender disparities in facial emotions were more severe than in Google Images. Furthermore, DALL·E 2 was more likely to present smiling women and women with their heads pitching down in female-dominated (vs. male-dominated) occupations, demonstrating the risk of amplifying presentational biases in the gender stereotypical direction. If widely adopted, DALL·E 2 generated images are likely to reinforce the stereotype of emotional and smiling women versus rational men (Renner & Masch, 2019) and further entrench the relationship between femininity and submission (Rose et al., 2012). Since our algorithm auditing study is descriptive in nature, we encourage future research to examine the "effects" of exposure to AI-generated gendered occupational images or the active participation in human-AI co-production of bias-laden media content. For example, repeated exposure to such AI-generated images might reinforce and further distort users' perceptions of occupational gender norms (Kay et al., 2015).

Lastly, given the potential for AI models to perpetuate various biases, recent works have raised questions about the ethical responsibilities of online information providers and ways to combat the reproduction of inequalities (Hofeditz et al., 2022; Quadflieg et al., 2022). Proposed solutions include legislative approaches, administrative regulations, non-discrimination by code or design, and ethics guidelines. Quadflieg and colleagues (2022) also called for individuals to act upon disobedience to resist the negative effects of AI power. However, few studies have integrated feminist epistemology into their analysis of AI gender bias. In her work examining past and current practices concerning feminist artificial intelligence, Toupin (2023) argues that feminist knowledge production within AI has been undervalued due to the prevailing rationalist paradigm of "male as norm." Furthermore, feminist perspectives were excluded from AI history, resulting in a masculinist and rationalist historical account. Toupin calls for rethinking AI with feminist epistemology and offering alternative narratives to challenge the status quo.

Additionally, Wellner and Rothman (2020) review four strategies for eliminating gender bias in AI: ignoring gender references, revealing algorithmic decision-making considerations, designing non-biased algorithms, and involving humans in the process. They advocate for increasing awareness of gender bias and making active efforts to eliminate it, based on the feminist understanding that visibility matters. In light of this thinking, it is important for developers and other stakeholders in the generative AI industry to take responsibility and incorporate feminist epistemology into their daily practices to reduce gender biases.

**Limitations**

This study has several limitations worth noting. First, we focused on the largest face multi-face images and followed the default classification threshold (50%) for face and gender detection, which may have resulted in losing nuances across different images. Second, the gender variable in our study is binary due to the limitations of the AWS *Rekognition* algorithm and the CPS census data. Future research should seek to expand gender categories to improve inclusivity. Moreover, both *Rekognition* algorithms and human coding relied on stereotypical conceptions of gender presentation so the study may have inadvertently reinforced binary gender stereotypes. Finally, although our study revealed gender biases in text-to-image generative AI from various perspectives, we did not conduct experimental studies to examine how exposure to such gendered images may affect people's perceptions of occupational gender norms and downstream beliefs, attitudinal, and behavioral consequences. We encourage future research to fill this gap. Moreover, future studies should continue algorithm auditing for gender biases in generative AI models, including DALLE·2, and test the effectiveness of potential de-biasing strategies such as technical, legal, administrative, and individual resistance approaches (Wellner & Rothman, 2020).

**Conclusion**

This study reveals that DALL·E 2, a popular image generative AI model, systematically underrepresents women in male-dominated occupations and overrepresents them in female-dominated jobs. Furthermore, DALL·E 2 images tend to portray more women than men with smiling faces and head pitching down, particularly in female-dominated (vs. male-dominated) occupations, reinforcing traditional gender stereotypes. Our computational algorithm auditing study thus demonstrates more severe representational and presentational biases in DALL·E 2 when compared to Google Images. These findings emphasize the importance of continuous monitoring and evaluation of gender biases in generative AI technologies. Future research should expand the scope of gender categories, examine the potential effects of exposure to gendered AI-generated images, and explore strategies to effectively mitigate gender biases in AI models.

**Data Availability**

Replication data and codes are available upon request.

**Notes**

1. See https://www.bls.gov/cps/tables.htm#annual: Labor Force Statistics from the Current Population Survey.
2. See https://www.census.gov/topics/population/age-and-sex/about.html and https://www2.census.gov/programs-surveys/cps/techdocs/questionnaires/Demographics.pdf: according to the Bureau of Census, the census uses the concept "sex" rather than "gender" in the questionnaire:
3. See https://serpapi.com: SerpAPI.
4. See https://aws.amazon.com/rekognition: Amazon AWS Rekognition.
5. See https://openai.com/research/dall-e-2-pre-training-mitigations

Collins, R. L. (2011). Content Analysis of Gender Roles in Media: Where Are We Now and Where Should We Go? *Sex Roles*, *64*(3), 290–298. https://doi.org/10.1007/s11199-010-9929-5

Coltrane, S., & Adams, M. (1997). Work–Family Imagery and Gender Stereotypes: Television and the Reproduction of Difference. *Journal of Vocational Behavior*, *50*(2), 323–347. https://doi.org/10.1006/jvbe.1996.1575

Davies, P., Quinn, D., & Gerhardstein Nader, R. (2002). Consuming Images: How Television Commercials That Elicit Stereotype Threat Can Restrain Women Academically and Professionally. *Personality and Social Psychology Bulletin*, *28*, 1615–1628. https://doi.org/10.1177/014616702237644

Döring, N., Reif, A., & Poeschl, S. (2016). How gender-stereotypical are selfies? A content analysis and comparison with magazine adverts. *Computers in Human Behavior*, *55*, 955–962. https://doi.org/10.1016/j.chb.2015.10.001

Entman, R. M. (2007). Framing Bias: Media in the Distribution of Power. *Journal of Communication*. https://doi.org/10.1111/j.1460-2466.2006.00336.x

Freelon, D. (2013). ReCal OIR: ordinal, interval, and ratio intercoder reliability as a web service. *International Journal of Internet Science*, *8*(1).

Gibbs, S. (2015, July 8). *Women less likely to be shown ads for high-paid jobs on Google, study shows | Google | The Guardian*. The Guardian. https://www.theguardian.com/technology/2015/jul/08/women-less-likely-ads-high-paid-jobs-google-study

Good, J. J., Woodzicka, J. A., & Wingfield, L. C. (2010). The Effects of Gender Stereotypic and Counter-Stereotypic Textbook Images on Science Performance. *The Journal of Social Psychology*, *150*(2), 132–147. https://doi.org/10.1080/00224540903366552

Grabe, M. E., & Bucy, E. P. (2009). *Image Bite Politics: News and the Visual Framing of Elections* (1st edition). Oxford University Press.

Grau, S., & Zotos, Y. (2016). Gender stereotypes in advertising: A review of current research. *International Journal of Advertising*, *35*, 761–770. https://doi.org/10.1080/02650487.2016.1203556

Hamilton, M. C., Anderson, D., Broaddus, M., & Young, K. (2006). Gender Stereotyping and Under-representation of Female Characters in 200 Popular Children's Picture Books: A
19

**Tables and Figures**
**Figure 1**
*Image Examples for Visual Gender Stereotypes*

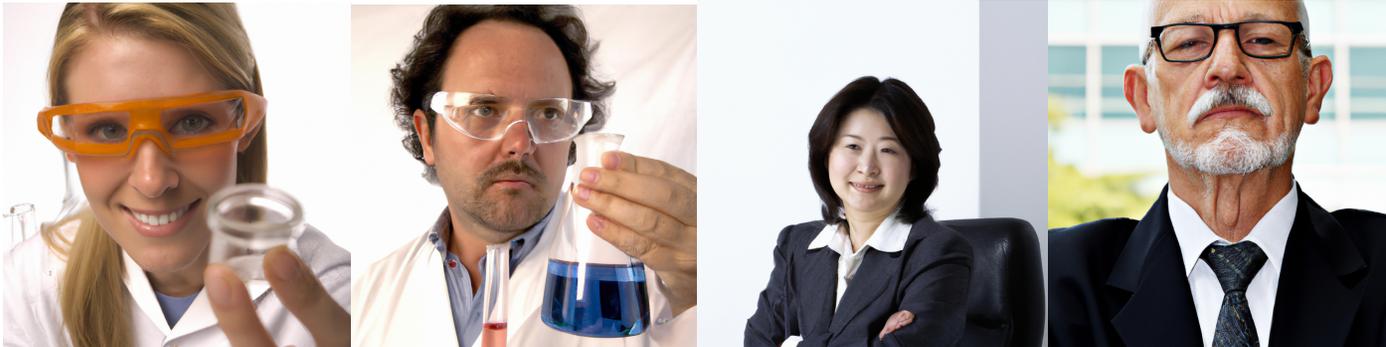

*Note.* All four image examples were collected from DALL·E 2. From left to right, the images are (a) a biological scientist detected as a woman with a smile; (b) a biological scientist detected as a man with calmness; (c) a chief executive officer detected as a woman with a lower pose pitch value; (d) a chief executive officer detected as a man with a higher pose pitch value.

**Figure 2**
*Paired Estimated Differences in Female Percentage between CPS Census Data, DALL·E 2, and Google Images*

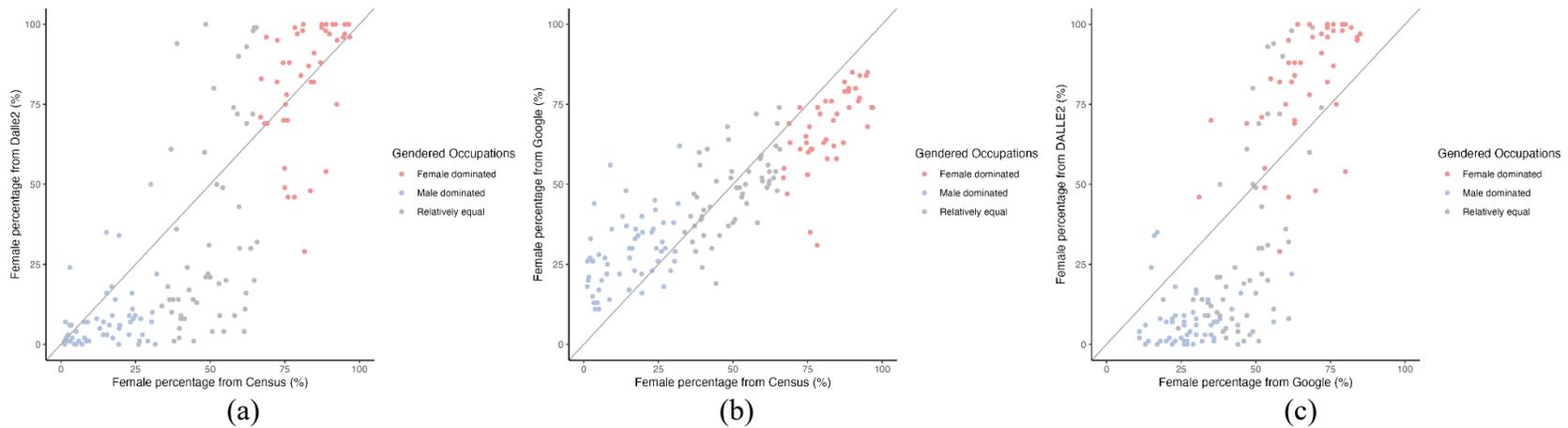

(a)            (b)            (c)

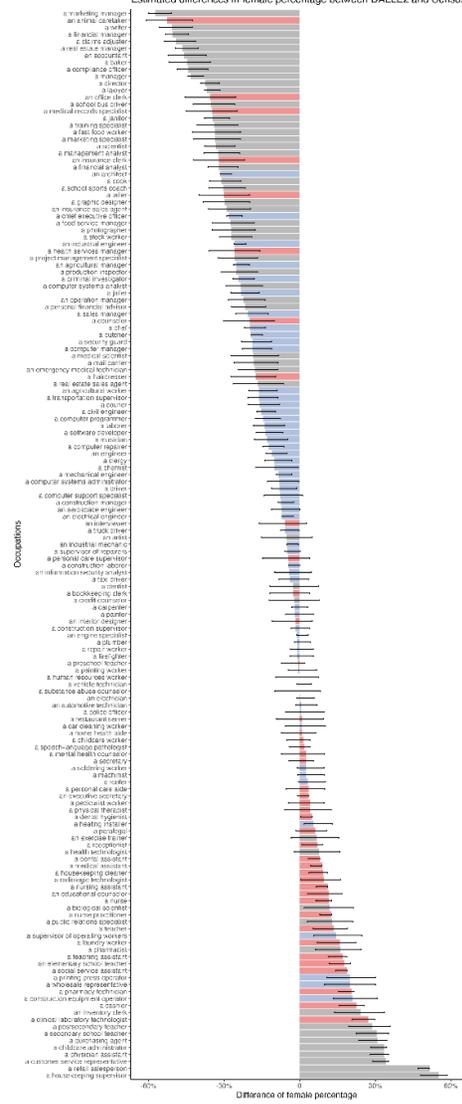
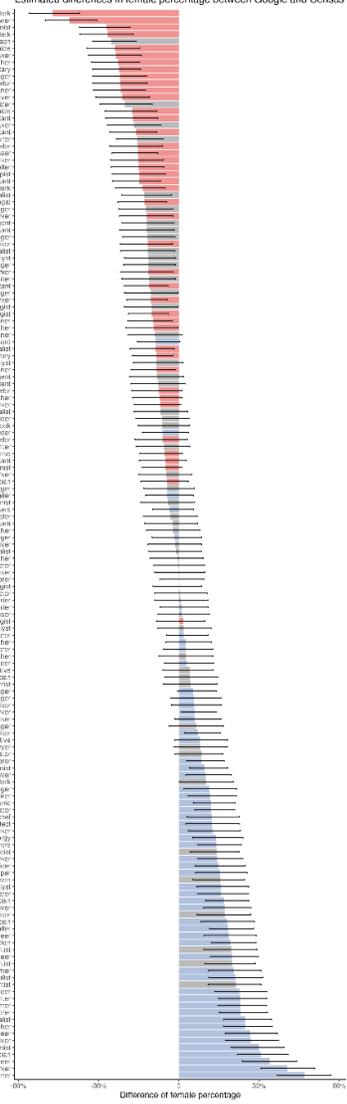
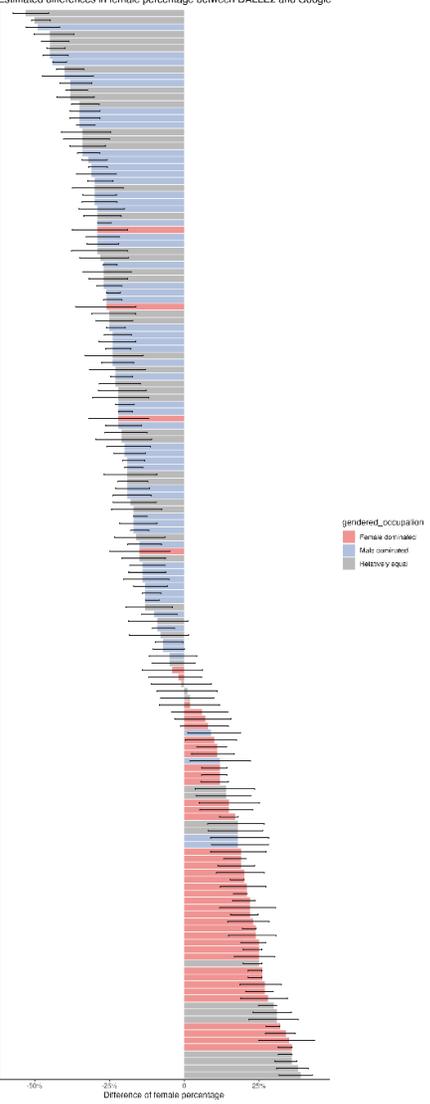

(d)                  (e)                  (f)

**Table 1**
*The Number of Occupations with Significantly Different Female Percentages*

|                      | Countering Gender Bias | Confirming Gender Bias | Total |
|----------------------|------------------------|------------------------|-------|
| DALL·E 2 v.s. Census | 20                     | 88                     | 108   |
| Google v.s. Census   | 79                     | 18                     | 97    |
| DALL·E 2 v.s. Google | 17                     | 122                    | 139   |

*Note.* "Counter Gender Bias" refers to the occupation

**Table 2**
*ANOVA Results for Smile*

|                                         | Model 1 |     |        | Model 2 |     |        |
|-----------------------------------------|---------|-----|--------|---------|-----|--------|
|                                         | $\chi^2$ | df  | p      | $\chi^2$ | df  | p      |
| Gender                                  | 317.71  | 1   | < .001 | 19.28   | 1   | < .001 |
| Source                                  | 477.58  | 1   | < .001 | 216.12  | 1   | < .001 |
| Occupation Category                     |         |     |        | 5.19    | 2   | .075   |
| Gender × Source                         | 13.84   | 1   | .001   | 1.54    | 1   | .214   |
| Gender × Occupation Category            |         |     |        | 27.64   | 2   | < .001 |
| Source × Occupation Category            |         |     |        | 2.29    | 2   | .319   |
| Gender × Source × Occupation Category   |         |     |        | 14.95   | 2   | .001   |

*Note.* $N$ = 30,600 (153 occupations) for Model 1; $N$ = 27,600 (138 occupations) for Model 2. For Model 2, occupations that had no female or male images in either source category (DALL·E 2 or Google) were excluded. By-occupation random intercepts were included in the models to account for the multi-level structure of the data.
† $p < .10$, * $p < .05$, ** $p < .01$, *** $p < .001$.

**Table 3**
*ANOVA Results for Calmness*

|  | Model 1 | | | Model 2 | | |
|---|---|---|---|---|---|---|
|  | F | df | p | F | df | p |
| Gender | 272.91 | 1, 27482.2 | < .001 | 33.06 | 1, 27559.1 | < .001 |
| Source | 64.45 | 1, 30544.8 | .001 | 16.25 | 1, 27485.1 | < .001 |
| Occupation Category |  |  |  | 12.58 | 2, 224.3 | < .001 |
| Gender × Source | 15.18 | 1, 30567.4 | .001 | 2.26 | 1, 27587.3 | .133 |
| Gender × Occupation Category |  |  |  | 1.84 | 2, 27570.6 | .159 |
| Source × Occupation Category |  |  |  | 3.53 | 2, 27552.2 | .029 |
| Gender × Source × Occupation Category |  |  |  | 0.11 | 2, 27580.7 | .899 |

*Note.* $N = 30,600$ (153 occupations) for Model 1; $N = 27,600$ (138 occupations) for Model 2. For Model 2, occupations that had no female or male images in either source category (DALLE or Google) were excluded. By-occupation random intercepts were included in the models to account for the multi-level structure of the data.
† $p < .10$, * $p < .05$, ** $p < .01$, *** $p < .001$.

**Table 4**

*ANOVA Results for Pose Pitch*

|  | Model 1 | | | Model 2 | | |
|---|---|---|---|---|---|---|
|  | F | df | p | F | df | p |
| Gender | 11.36 | 1, 24035.9 | .001 | 0.66 | 1, 27564.3 | .416 |
| Source | 163.68 | 1, 30572.7 | < .001 | 34.77 | 1, 27487.5 | < .001 |
| Occupation Category |  |  |  | 3.97 | 2, 232.3 | .020 |
| Gender × Source | 0.60 | 1, 30420.6 | .440 | 1.01 | 1, 27588 | .315 |
| Gender × Occupation Category |  |  |  | 1.35 | 2, 27575.2 | .259 |
| Source × Occupation Category |  |  |  | 9.07 | 2, 27557 | < .001 |
| Gender × Source × Occupation Category |  |  |  | 1.72 | 2, 27575.4 | .179 |

*Note.* $N = 30{,}600$ (153 occupations) for Model 1; $N = 27{,}600$ (138 occupations) for Model 2. For Model 2, occupations that had no female or male images in either source category (DALL·E 2 or Google) were excluded. By-occupation random intercepts were included in the models to account for the multi-level structure of the data.
† $p < .10$, * $p < .05$, ** $p < .01$, *** $p < .001$.

**Figure 3**

*Probability of Smile*

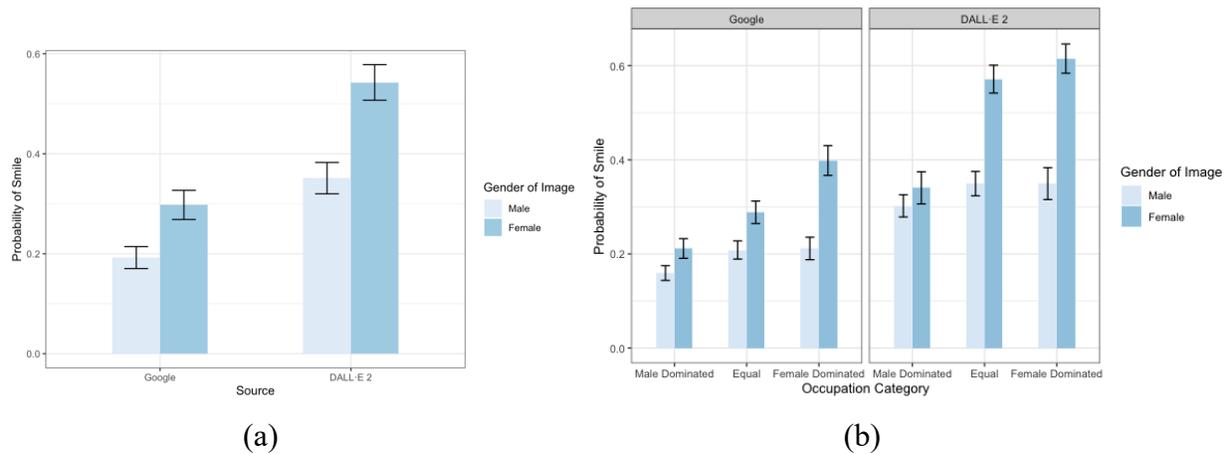

(a)  (b)

*Note.* Panel (a) presents results from Table 2, Model 1; panel (b) presents results from Table 2, Model 2. Error bars are 95% confidence intervals.

**Figure 4**

*Calmness*

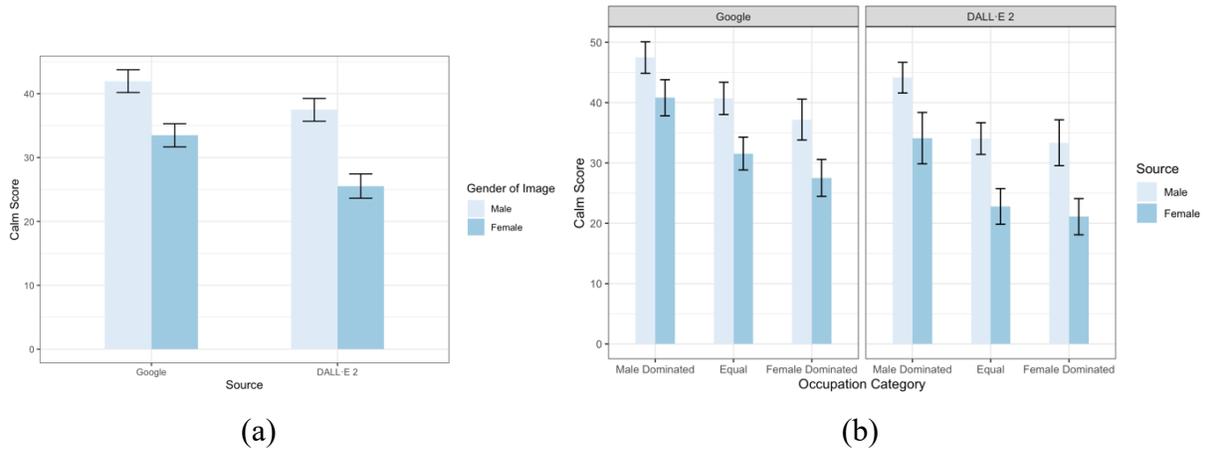

(a)            (b)

*Note*. Panel (a) presents results from Table 3, Model 1; panel (b) presents results from Table 3, Model 2. Error bars are 95% confidence intervals.

**Figure 5**

*Pose Pitch*

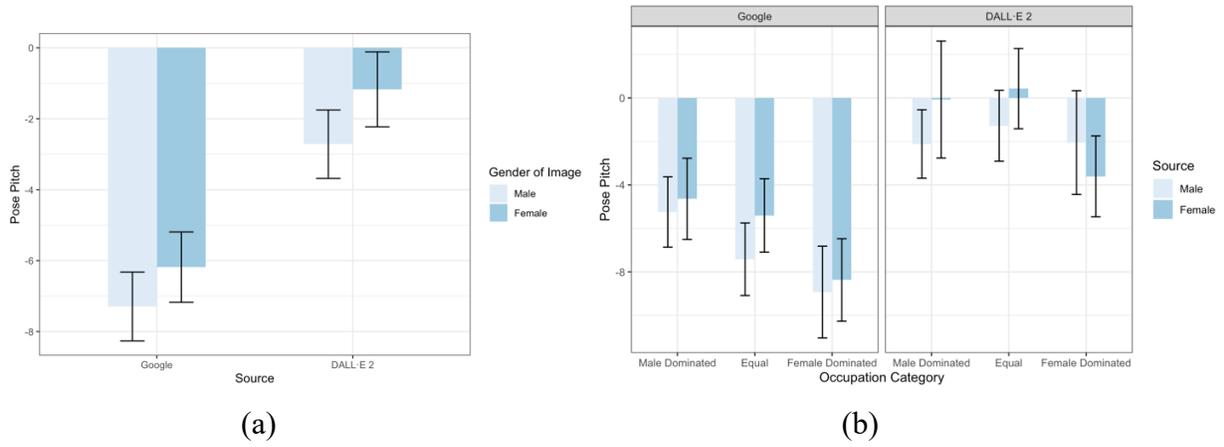

(a)  (b)

*Note*. Panel (a) presents results from Table 4, Model 1; panel (b) presents results from Table 4, Model 2. Error bars are 95% confidence intervals.

# Appendix A

**Table A1**

*Generalized Linear Mixed Model Results for Smile*

|  | Model 1 | Model 2 |
|---|---|---|
|  | OR [95% CI] | OR [95% CI] |
| Gender [a] | 2.19 [2.01, 2.39] | 1.19 [0.95, 1.49] |
| Source [b] | 0.44 [0.41, 0.47] | 0.44 [0.39, 0.49] |
| Relatively Equal [c] |  | 1.24 [0.91, 1.70] |
| Female Dominated [c] |  | 1.24 [0.86, 1.79] |
| Gender × Source | 0.81 [0.73, 0.91] | 1.19 [0.91, 1.56] |
| Gender × Relatively Equal |  | 2.08 [1.60, 2.71] |
| Gender × Female Dominated |  | 2.50 [1.87, 3.32] |
| Source × Relatively Equal |  | 1.12 [0.95, 1.32] |
| Source × Female Dominated |  | 1.14 [0.90, 1.44] |
| Gender × Source × Relatively Equal |  | 0.52 [0.38, 0.73] |
| Gender × Source × Female Dominated |  | 0.70 [0.49, 1.00] |

*Note.* $N = 30,600$ (153 occupations) for Model 1; $N = 27,600$ (138 occupations) for Model 2. For Model 2, occupations that had no female or male images in either source category (DALL·E 2 or Google) were excluded. By-occupation random intercepts were included in the models to account for the multi-level structure of the data. Cell entries are unstandardized coefficients with standard errors in parentheses. Statistical significance for each coefficient was tested using the estimated standard errors (Wald test). OR indicates odds ratio.
[a] Female = 1, Male = 0. [b] Google = 1, DALL·E 2 = 0. [c] The occupation category variable includes two dummy coded variables. The reference group is Male Dominated.

**Table A2**
*Linear Mixed Model Results for Calmness*

|  | Model 1 | Model 2 |
| --- | --- | --- |
| Gender [a] | −11.94*** (0.72) | −10.02*** (1.91) |
| Source [b] | 4.50*** (0.56) | 3.34*** (0.83) |
| Relatively Equal [c] |  | −10.09*** (1.87) |
| Female Dominated [c] |  | −10.79*** (2.33) |
| Gender × Source | 3.45*** (0.89) | 3.36 (2.23) |
| Gender × Relatively Equal |  | −1.23 (2.26) |
| Gender × Female Dominated |  | −2.23 (2.47) |
| Source × Relatively Equal |  | 3.32** (1.27) |
| Source × Female Dominated |  | 0.50 (1.90) |
| Gender × Source × Relatively Equal |  | −1.25 (2.72) |
| Gender × Source × Female Dominated |  | −0.76 (3.00) |

*Note.* $N = 30{,}600$ (153 occupations) for Model 1; $N = 27{,}600$ (138 occupations) for Model 2. For Model 2, occupations that had no female or male images in either source category (DALL·E 2 or Google) were excluded. By-occupation random intercepts were included in the models to account for the multi-level structure of the data. Cell entries are unstandardized coefficients with standard errors in parentheses. Statistical significance for each coefficient was tested using the estimated standard errors (Wald test).
[a] Female = 1, Male = 0. [b] Google = 1, DALL·E 2 = 0. [c] The occupation category variable includes two dummy coded variables. The reference group is Male Dominated.
† $p < .10$, * $p < .05$, ** $p < .01$, *** $p < .001$.

**Table A3**
*Linear Mixed Model Results for Pose Pitch*

|  | Model 1 | Model 2 |
|---|---|---|
| Gender [a] | 1.55*** (0.46) | 2.04† (1.22) |
| Source [b] | –4.58*** (0.36) | –3.13*** (0.53) |
| Relatively Equal [c] |  | 0.84 (1.16) |
| Female Dominated [c] |  | 0.06 (1.46) |
| Gender × Source | –0.44 (0.56) | –1.44 (1.43) |
| Gender × Relatively Equal |  | –0.34 (1.45) |
| Gender × Female Dominated |  | –3.60* (1.58) |
| Source × Relatively Equal |  | –3.01*** (0.81) |
| Source × Female Dominated |  | –3.75** (1.22) |
| Gender × Source × Relatively Equal |  | 1.75 (1.75) |
| Gender × Source × Female Dominated |  | 3.55† (1.92) |

*Note.* $N = 30{,}600$ (153 occupations) for Model 1; $N = 27{,}600$ (138 occupations) for Model 2. For Model 2, occupations that had no female or male images in either source category (DALL·E 2 or Google) were excluded. By-occupation random intercepts were included in the models to account for the multi-level structure of the data. Cell entries are unstandardized coefficients with standard errors in parentheses. Statistical significance for each coefficient was tested using the estimated standard errors (Wald test).
[a] Female = 1, Male = 0. [b] Google = 1, DALL·E 2 = 0. [c] The occupation category variable includes two dummy coded variables. The reference group is Male Dominated.
† $p < .10$, * $p < .05$, ** $p < .01$, *** $p < .001$.